\documentclass[12pt]{elsarticle} 
\usepackage[a4paper, margin=3cm]{geometry} 
\usepackage{setspace} 
\usepackage{amsmath, amssymb, amsfonts}

\usepackage{algorithm}
\usepackage{algpseudocode}

\usepackage{graphicx}
\usepackage{textcomp}
\usepackage{xcolor}
\usepackage{caption}
\usepackage{tabularx}
\usepackage{array}
\usepackage{comment}
\usepackage{float}
\usepackage{booktabs}
\usepackage{enumitem}
\usepackage{hyperref} 
\usepackage{makecell}
\usepackage{subcaption}
\usepackage{algpseudocode}
\usepackage[font=normalsize]{caption}

\biboptions{numbers,sort&compress}
\makeatletter
\renewcommand\NAT@open{[} 
\renewcommand\NAT@close{]}
\makeatother

\begin{document}

\begin{frontmatter}

\title{Predictive Modeling of Power Outages during Extreme Events: Integrating Weather and Socio-Economic Factors}

\author{Nina Fatehi, Antar Kumar Biswas and Masoud H. Nazari*}
\cortext[cor1]{Corresponding author: Masoud H. Nazari (email: masoud.nazari@wayne.edu)}

\address{Department of Electrical and Computer Engineering, Wayne State University, Detroit, Michigan.}

\begin{abstract}
This paper presents a novel learning-based framework for predicting power outages caused by extreme events. The proposed approach specifically targets low-probability, high-consequence outage scenarios and leverages a comprehensive set of features derived from publicly available data sources. We integrate EAGLE-I outage records (2014–2024) with weather, socio-economic, infrastructure, and seasonal event data. Incorporating social and demographic indicators reveals underlying patterns of community vulnerability and provides a clearer understanding of outage risk during extreme conditions.
Four machine learning models—Random Forest (RF), Graph Neural Network (GNN), Adaptive Boosting (AdaBoost), and Long Short-Term Memory (LSTM) —are evaluated. 
Experimental validation is performed on a large-scale dataset covering counties in the lower peninsula of Michigan. 
Among all models tested, the LSTM network achieves higher accuracy.
\end{abstract}


\end{frontmatter}

\section{Introduction}

High-Impact Low-Probability (HILP) events—such as floods, hurricanes, and winter storms—pose severe threats to power systems reliability and resilience. 
\cite{xu2023review}. These events can cause large-scale blackouts and economic losses, as seen in Hurricane Sandy (2012), Irma (2017), and the 2021 Texas winter storm \cite{buzzfeed2022, texastribune2021, daeli2022power}. Several studies have explored a wide range of modeling approaches to understand and predict the relationship between extreme weather conditions and outage occurrence.

For example, Eskandarpour et al. \cite{R13} predicted component damage prior to hurricanes by establishing decision boundaries using logistic regression. Spatial–temporal poisson regression has also been applied to model transmission line failure probabilities during severe weather events \cite{yang2019}. Although these methods offer clear interpretability, their linear assumptions limit their ability to capture the complex multi-variable interactions.

ML models provide an effective framework to manage complex non-linear relationships and process substantial data volumes in real-time \cite{N30}.
These models outperform traditional statistical approaches due to their ability to model nonlinearities and integrate diverse datasets, including weather, infrastructure, and socio-economic indicators. Classification-based ML studies such as \cite{N3, eskandarpour2017leveraging} used models like SVM and boosted classifiers to determine outage occurrence or component damage. Regression-based studies have used tree-based ensembles such as Decision Trees, RF, AdaBoost, and XGBoost to predict outage counts, durations, or affected customers \cite{N1, cruz, N41}. Deep learning approaches further extend predictive capability by capturing temporal and sequential patterns in outage and weather data. Arif et al. \cite{N4} employed deep neural networks to estimate restoration times, and recurrent architectures such as recurrent neural networks (RNN) to model temporal outage progression under varying environmental conditions. These models offer strong predictive performance but require large, high-quality datasets. Some studies examined on the outages driven by routine weather or equipment failures. Jaech et al. \cite{R14} applied RNN-based models with environmental and operational data to predict outage duration in Seattle, while Sun et al. \cite{R12} used AdaBoost to estimate outage counts in Kansas. Sarwat et al. \cite{sarwat2016} showed that day-to-day weather conditions influence power outage. Our work focuses on HILP events that generate large-scale disruptions and require fundamentally different modeling considerations.

The state-of-the-art studies often encounter challenges that can compromise the accuracy of their analyses.
A major limitation is the scarcity of historical data on extreme events and the associated power outages. In particular, during such events, several monitoring devices may fail, resulting in incomplete and imbalanced datasets.
In \cite{N26, N14, N17}, Synthetic Minority Oversampling Technique (SMOTE) has been used to address data imbalance in classification tasks in power outage studies. However, SMOTE is effective for categorical targets, it is less productive for HILP outage prediction, which is framed as a regression problem.  
Another limitations arises when models are trained on a single extreme event, as they exhibit limited adaptability to concurrent incidents with complex, multi-variable interactions \cite{N15}. As shown in \cite{N16}, combining data from two winter storm events instead of one improves power outage prediction. 
Furthermore, socio-economic features are often missing in the previous models. As demonstrated in Cruz et al. \cite{cruz}, socio-economic indicators such as income levels and housing characteristics provide important context for understanding community vulnerability. Counties with lower income or higher unemployment tend to have fewer resources for preparedness, which can increase outage severity and prolong restoration times. Similar observations have been reported in other studies \cite{N4, R14}, where result shows that socio-economic factors contribute to prolonged outage durations and higher customer outage number. Incorporating socio-economic variables therefore enables a more comprehensive understanding of outage dynamics, supports improved prediction performance for HILP events.

In our prior work \cite{N17}, we developed ML models to predict power outage probabilities, investigating the impact of socio-economic and infrastructure factors on non-HILP outage events. This paper builds on those findings and proposes enhanced methodologies to predict power outages due to extreme weather events. The contributions of this paper are outlined as follows: 

    1) This model combines dynamic data such as weather, seasonal indicators, historical power outages with static data such as socio-economic, and infrastructure data to capture the complex factors influencing power outages. This comprehensive integration significantly enhances the predictive performance of the model.
   
   2) This work introduces the application of the Synthetic Minority Over-sampling Technique for Regression (SMOGN) and K-Nearest Neighbors (KNN) methods to mitigate data imbalance and missing data challenges associated with HILP events. By adapting these techniques to the outage prediction context, the proposed framework effectively addresses data scarcity and improves the accuracy and robustness of HILP event prediction.

   3) We evaluate and compare multiple ML and deep learning (DL) models for extreme event outage prediction using integrated static and dynamic data. This makes our model highly suited for real-world power outage forecasting.
   
   4) We train our approach on a
large-scale power outage dataset across several Michigan counties. The dataset includes historical extreme events from 2014 to 2024, including the flooding event in Wayne county in June, 2020. This model can be generalized to diverse geographic regions and socio-economic conditions.

The remainder of the paper is structured as follows: Section II presents the data description and preparation. In section III, the methodology is discussed. Experimental setup is presented in section IV. The experimental results and discussion are provided in section V. Finally, the conclusion  is drawn in section VI.

\section{Data Description and Preparation}
\subsection{Historical Outage Data}
\begin{figure}[t]
\centering
\includegraphics[scale=0.35]{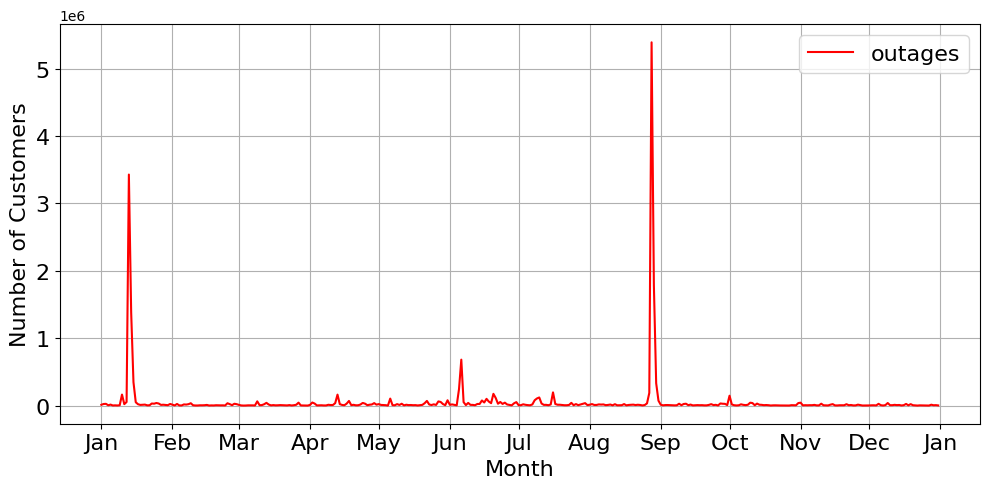}
\caption{Number of customers outage in Wayne County during 2024 based on EAGLE-1 dataset.}
\label{Wayne Outage}
\end{figure}

In this study, the EAGLE-I dataset is utilized to analyze severe power outages caused by extreme weather events. 
%
The EAGLE-I dataset \cite{N25}, collected and managed by Oak Ridge National Laboratory (ORNL) on behalf of DOE, spans from November 2014 to December 2024. This dataset, updated every 15 minutes, aggregates data from utility companies' public outage maps using an ETL (Extract, Transform, Load) process to estimate the number of customers affected by power outages in each county. For this analysis, we focus on the outage data relevant to Michigan. Fig. \ref{Wayne Outage} illustrates the daily number of customer outages in Wayne County throughout 2024. In this context, a 'customer' is defined as any entity that purchases energy from a utility under a specific tariff, rather than individual persons.

\subsection{Historical Weather Data}
Historical hourly weather data was obtained from Open-Meteo, an open-source weather API that provides high-resolution meteorological information on an hourly basis \footnote[3]{https:https://open-meteo.com/}. This data source serves as the dynamic variable of the model. For instance, weather data offers insights into conditions that could lead to outages, such as wind speed and temperature. We have chosen 8 features including air temperature (F), precipitation (inch), wind speed (km/h), wind gusts (km/h), shortwave radiation ($W/m^2$), relative humidity (\%), cloud cover (\%), and surface pressure (hPa). 
The historical weather data often has missing data, particularly during HILP events.  
To address this issue, KNN imputation method has been applied \cite{biswas5151316data}. For a given county, the \( k \) closest counties are determined by calculating the Euclidean distance between county. Then, impute the missing data by averaging the selected nearest neighbors, identified by the KNN algorithm as follows in Algorithm \ref{alg:knn}.

\begin{algorithm}[h!]
\caption{Algorithm to Estimate Missing Data}
\label{alg:knn}
\begin{algorithmic}[0.5]
\State \textbf{Input:} 
\State \hspace{2mm} \text{Set of counties }$\mathcal{C}=\{C_1,C_2,\dots,C_N\}$
\State \hspace{2mm} \text{Number of nearest counties} $k$ (set to $k=5$)
\State \hspace{2mm} \text{Geographic coordinates} $(x_i,y_i)$ of county $C_i$
\State \hspace{2mm} $D_{ij}$ \text{ (Euclidean distance between $C_i$ and $C_j$)}
\vspace{5pt}

\State \textbf{Procedure:}
\For{each county $C_i \in \mathcal{C}$}
    \State Initialize distance list $\mathcal{D}_i \leftarrow \emptyset$
    \For{each county $C_j \in \mathcal{C} (i\not= j)$}
            \State Compute distance:
            \[
            D_{ij} = \sqrt{(x_i - x_j)^2 + (y_i - y_j)^2}
            \]
            \State Store $(C_j, D_{ij})$ to $\mathcal{D}_i$
    \EndFor
    \State Sort $\mathcal{D}_i$ in ascending order of $D_{ij}$
    \State Select the first $k$ counties from $\mathcal{D}_i$
    \State Compute the average value of the $k$ neighbors
\EndFor
\vspace{5pt}

\State \textbf{Output:} Impute the missing data
\end{algorithmic}
\end{algorithm}

\begin{figure}[h]
\centering
\includegraphics[scale=0.4]{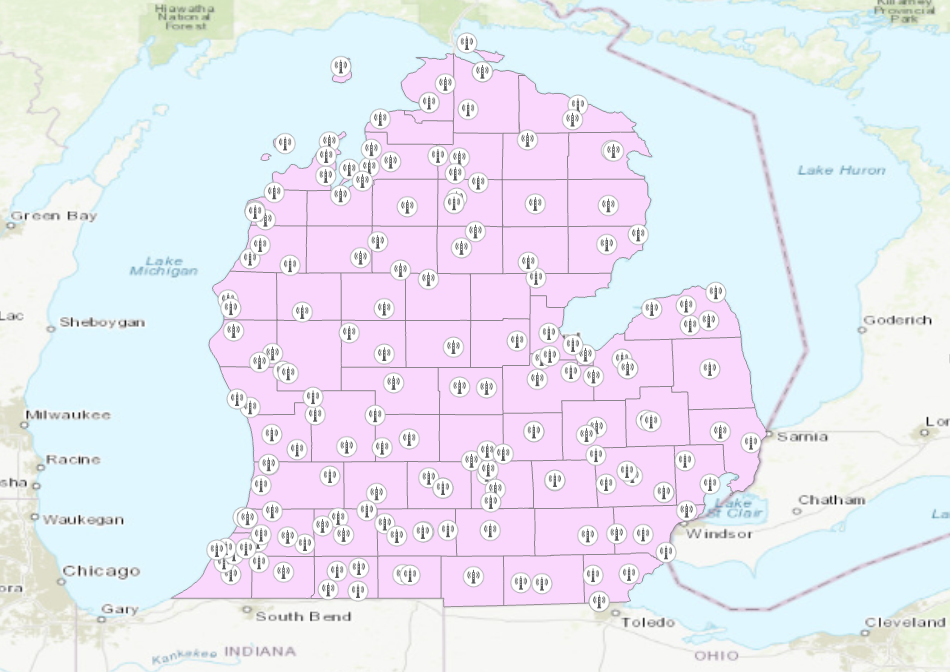}
\caption{Weather stations locations in Lower Peninsula of Michigan.}
\label{weather stations_}
\end{figure}

\subsection{Socio-economic  Data}
Socio-economic indicators provide meaningful insights into community vulnerability and resilience. This hidden knowledge is instrumental in understanding factors that are not immediately obvious from basic data analysis. 
For this study, socio-economic indicators were obtained from the U.S. Census Bureau, including the distribution of the year when residential structures were built (2021 estimate), the average household income (2021 estimate), and the county-level unemployment rate (2021 estimate). The selected data was retrieved from the U.S. Census Bureau’s 2021 American Community Survey (ACS) 5-Year Estimates via the Census API. \footnote[4]{https://api.census.gov/data/2021/acs/acs5}

The distribution of the year when residential structures were built reflects the age and durability of the electric infrastructure. Average household income and unemployment capture the economic capacity and stress of a community, influencing both infrastructure investment and the speed of recovery from extreme weather events. Together, these factors complement power infrastructure data by capturing socio-economic conditions that influence outage occurrence and restoration.

\subsection{Power Infrastructure Data}

\begin{figure}[h]
\centering
\includegraphics[scale=0.5]{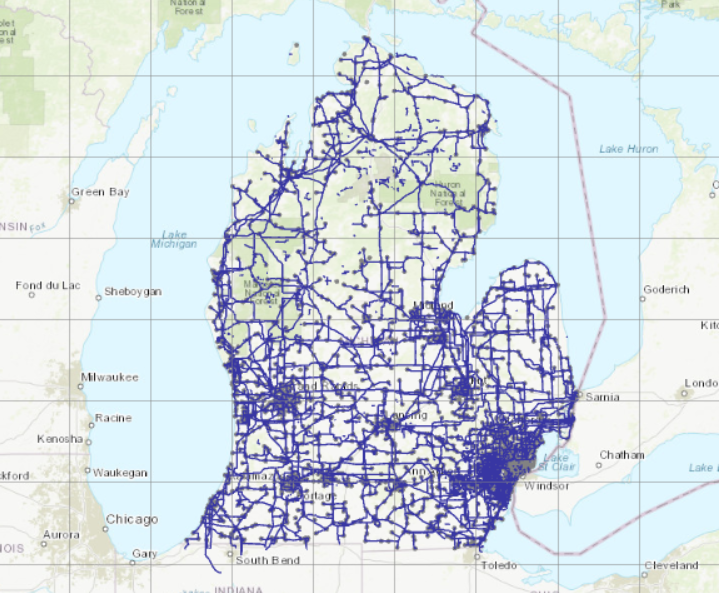}
\caption{Power infrastructure network of 20 counties in Lower Michigan.}
\label{power_infra}
\end{figure}

The infrastructure features, including the number of poles, towers, substations, transformers, and lines for each county, are collected from the Open Street Map dataset \cite{N19} as shown in Fig. \ref{power_infra}. 
Power infrastructure data provides valuable insights into the characteristics of each county. Given that the detailed topology of the power distribution network is typically not publicly available, data from the Open Street Map can serve as a useful approximation of the actual topology for research purposes. Furthermore, they provide information on how extensively the system might be affected by extreme conditions such as high winds \cite{N6}.

For this study, the total number of each feature within a county is aggregated and then counted to provide a comprehensive overview of the area's characteristics.
Then, we perform column-wise normalization, where we divide the total number of each type of infrastructure in a county by the sum of that feature across the entire state. This method allows us to determine which county has a higher proportion of a particular infrastructure component, such as towers.

\section{Methodology}
\begin{figure*}[h]
    \centering
    \includegraphics[width=0.9\textwidth]{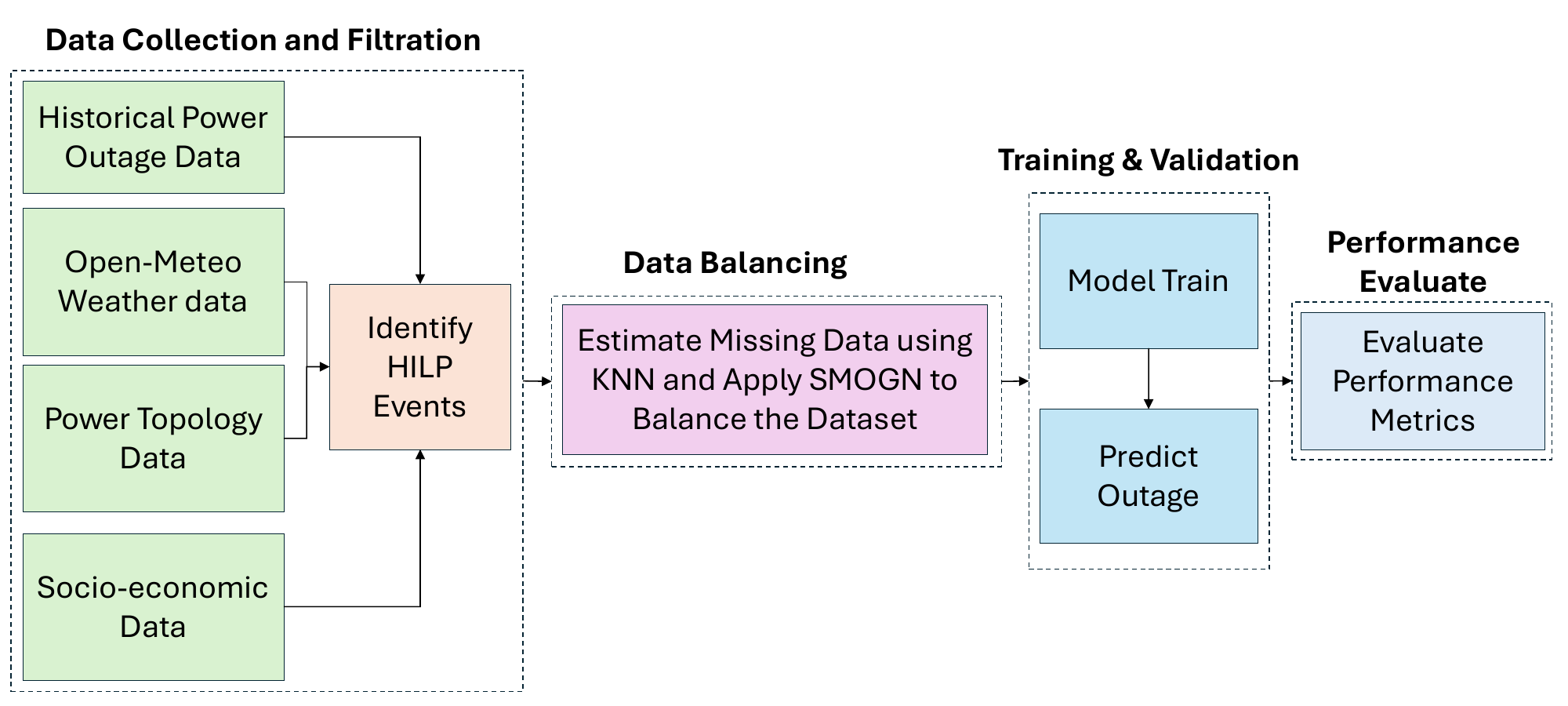}
    \caption{The proposed learning-based framework for power outage prediction.}
    \label{Framework1}
\end{figure*}

The proposed multi-step county-level power outage prediction algorithm due to extreme weather events is illustrated in Fig. \ref{Framework1}. 
The algorithm follows a four-step process. First, data preparation is performed, where different input features are preprocessed and integrated into a unified dataset. Second, the integrated dataset is rebalanced to reduce the impact of data imbalance and missing data. Third, multiple machine learning models are trained on the data to capture the complex relationships between weather, socio-economic, and infrastructure factors and outage occurrence. Finally, the performance of the models is evaluated using standard metrics. 
In the following, we discuss the last three steps, and then the model formulation is presented.

\subsection{Identifying Extreme Weather Event Data}

\begin{figure}[h!]
\centerline{\includegraphics[scale=0.35]{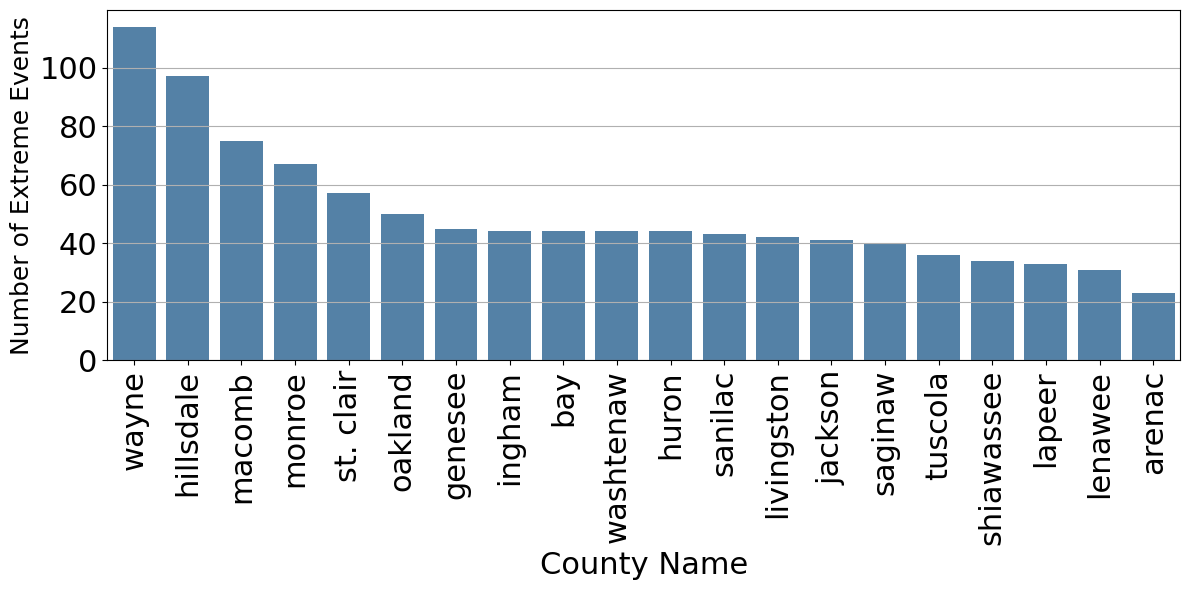}}
\caption{Number of Extreme Events by County (2014–2024).}
\label{event_count}
\end{figure}

The outage and weather dataset spans the period from 1 November 2014 to 30 December 2024. 
Within this time frame, a total of 236 weather-related extreme outage events were identified from the National Centers for Environmental Information (NOAA) \footnote[5]{https://www.ncdc.noaa.gov/stormevents/}. Fig. \ref{event_count} illustrates the distribution of extreme weather events across 20 Michigan counties, showing that Wayne and Hillsdale counties experienced the highest number of events. 

This distribution shows that extreme events are sparse. To address this limitation, seasonal information aligned with the timing of distinct extreme weather patterns is incorporated. Since HILP events are characterized by large outage magnitudes, customer outage counts are used directly as the severity indicator. Two complementary indicators are defined to identify HILP outage conditions. Let $y(c,t)$ denote the number of customers without power in county $c$ at time $t$, as reported by the EAGLE-I dataset. Since HILP events are characterized by large outage magnitudes, customer outage counts are used directly as the severity indicator.

\subsubsection{Identification of HILP Outage Events}

Two complementary indicators are defined to identify HILP outage conditions.

First, a weather-based extreme indicator is defined as
\begin{equation}
I^{\mathrm{WX}}_{c,t} =
\begin{cases}
1, & \text{if a NOAA-reported extreme weather occurs in county } c \text{ at time } t, \\
0, & \text{otherwise}.
\end{cases}
\label{eq:wx_indicator}
\end{equation}

Second, an extreme outage indicator is defined using the empirical distribution of customer outage counts during extreme weather periods. Let $Q_{\alpha}$ denote the $\alpha$-quantile of $y(c,t)$. An outage is considered high-consequence if
\begin{equation}
I^{\mathrm{HC}}_{c,t} =
\begin{cases}
1, & y(c,t) \ge Q_{\alpha}, \\
0, & \text{otherwise},
\end{cases}
\label{eq:hc_indicator}
\end{equation}
where $\alpha = 0.7$ is used in this study, corresponding to the top 30\% of outage magnitudes observed during extreme weather conditions.

The set of HILP seed events is therefore defined as
\begin{equation}
\mathcal{E}_0 =
\left\{ (c,t) \mid I^{\mathrm{WX}}_{c,t} \cdot I^{\mathrm{HC}}_{c,t} = 1 \right\}.
\label{eq:HILP_seeds}
\end{equation}

These seed events represent extreme outage and serve as reference points for identifying meteorologically similar extreme conditions.

\subsubsection{Seasonal Analog Selection Using Weather Similarity}

For each HILP seed event $(c,t_{\mathrm{ex}}) \in \mathcal{E}_0$, a set of candidate analog periods is selected to preserve seasonal consistency. Let $m(t)$ denote the calendar month associated with time $t$. The seasonal candidate set is defined as
\begin{equation}
\mathcal{R}(c,t_{\mathrm{ex}}) =
\left\{ (c,t) \;\middle|\;
m(t) \in \left\{ m(t_{\mathrm{ex}})-1,\; m(t_{\mathrm{ex}}),\; m(t_{\mathrm{ex}})+1 \right\}
\right\}.
\label{eq:seasonal_set}
\end{equation}

For each county and time, an aggregated weather feature vector is constructed as
\begin{equation}
\mathbf{x}_{c,t} =
\big[
T_{c,t},\,
\mathrm{Prec}_{c,t},\,
WS_{c,t},\,
WG_{c,t},\,
SWR_{c,t},\,
RH_{c,t},\,
CC_{c,t},\,
SP_{c,t}
\big]^{\top},
\label{eq:weather_vector}
\end{equation}
where the variables represent daily or hourly aggregates of temperature, precipitation, wind speed, wind gust, shortwave radiation, relative humidity, cloud cover, and surface pressure.

To ensure balanced comparison across heterogeneous weather variables, feature-wise standardization is applied:
\begin{equation}
\tilde{\mathbf{x}}_{c,t} =
\frac{\mathbf{x}_{c,t} - {\mu}}{{\sigma}},
\label{eq:standardization}
\end{equation}
where ${\mu}$ and ${\sigma}$ denote the mean and standard deviation vectors computed over the full weather dataset.

Weather similarity between a seed event $t_{\mathrm{ex}}$ and a candidate time $t$ is quantified using the Euclidean distance in standardized feature space:
\begin{equation}
\delta(c,t_{\mathrm{ex}},t)
=
\left\|
\tilde{\mathbf{x}}_{c,t_{\mathrm{ex}}}
-
\tilde{\mathbf{x}}_{c,t}
\right\|_2 .
\label{eq:distance}
\end{equation}

For each seed event, the $K$ most similar analog periods are selected as
\begin{equation}
\mathcal{A}(c,t_{\mathrm{ex}})
=
\operatorname*{arg\,min}_{\substack{
t \in \mathcal{R}(c,t_{\mathrm{ex}}) \\
|\mathcal{A}| = K
}}
\delta(c,t_{\mathrm{ex}},t).
\label{eq:analog_selection}
\end{equation}

The final extreme-event dataset used for training is constructed as
\begin{equation}
\mathcal{E}
=
\mathcal{E}_0
\;\cup\;
\bigcup_{(c,t_{\mathrm{ex}})\in\mathcal{E}_0}
\mathcal{A}(c,t_{\mathrm{ex}}).
\label{eq:final_extreme_set}
\end{equation}

This procedure expands the HILP conditions by incorporating meteorologically similar extreme-weather periods while preserving seasonal characteristics.

\begin{algorithm}[h!]
\caption{Data Rebalancing for High-Impact Customer Outages}
\label{alg:elsavior}
\begin{algorithmic}[1]
\State \textbf{Input:}
\State \hspace{5mm} $\mathcal{T} \gets$ Hourly training set with feature matrix $\mathbf{Z}$ and target $q$
\State \hspace{5mm} $\tau \gets$ Outage-impact threshold
\State \hspace{5mm} $K \gets$ Number of nearest neighbors
\State \hspace{5mm} $\rho_o \gets$ Over-sampling rate
\State \hspace{5mm} $\rho_u \gets$ Under-sampling rate
\vspace{5pt}

\State \textbf{Partition:}
\State \hspace{5mm} $\mathcal{C}_{\text{hi}} \gets \{(\mathbf{z}_j, q_j) \in \mathcal{T} \mid q_j \ge \tau\}$ \hfill (High-impact cases)
\State \hspace{5mm} $\mathcal{C}_{\text{lo}} \gets \{(\mathbf{z}_j, q_j) \in \mathcal{T} \mid q_j < \tau\}$ \hfill (Low-impact cases)
\State \hspace{5mm} $\mathcal{T}_{\text{bal}} \gets \emptyset$
\vspace{5pt}

\State \textbf{/* Synthetic generation for high-impact outages */}
\For{$(\mathbf{z}_r, q_r) \in \mathcal{C}_{\text{hi}}$}
    \State $N_{\text{syn}} \gets \rho_o$
    \State $\mathcal{N}_r \gets$ $K$ nearest neighbors of $(\mathbf{z}_r, q_r)$ in $\mathcal{C}_{\text{hi}}$
    \State $\Delta_r \gets$ pairwise distances between $(\mathbf{z}_r, q_r)$ and $\mathcal{N}_r$
    \State $\lambda_r \gets 0.5 \times \text{median}(\Delta_r)$
    \For{$m = 1$ to $N_{\text{syn}}$}
        \State $(\mathbf{z}_m, q_m) \gets$ randomly selected neighbor from $\mathcal{N}_r$
        \If{$\mathrm{dist}\big((\mathbf{z}_m, q_m), (\mathbf{z}_r, q_r)\big) < \lambda_r$}
            \State $(\mathbf{z}^{*}, q^{*}) \gets$ SMOTER interpolation between $(\mathbf{z}_r, q_r)$ and $(\mathbf{z}_m, q_m)$
        \Else
            \State $\mathbf{z}^{*} \gets \mathbf{z}_r + \boldsymbol{\eta}$ \hfill (Gaussian perturbation)
            \State $q^{*} \gets q_r$
        \EndIf
        \State Append $(\mathbf{z}^{*}, q^{*})$ to $\mathcal{T}_{\text{bal}}$
    \EndFor
\EndFor
\vspace{5pt}

\State \textbf{/* Down-sampling low-impact outages */}
\State $\mathcal{C}_{\text{lo}}^{\downarrow} \gets$ randomly sample $|\mathcal{C}_{\text{lo}}| \times \rho_u$ elements from $\mathcal{C}_{\text{lo}}$
\State Store $\mathcal{C}_{\text{lo}}^{\downarrow}$ to $\mathcal{T}_{\text{bal}}$
\vspace{5pt}

\State \textbf{Output:} Balanced training dataset $\mathcal{T}_{\text{bal}}$
\end{algorithmic}
\end{algorithm}

\subsection{Input Data Balancing}

A key challenge in predicting HILP outages is the imbalance in outage data, where small-scale events dominate and large-scale extreme events are minority. This imbalance causes ML models to favor frequent low-impact cases, leading to poor performance on the critical but infrequent HILP events. In order to balance the dataset, we apply SMOGN, which has been introduced to address the issue of imbalanced datasets in regression tasks \cite{N20}.  It integrates two over-sampling technique, SMOTER and SMOTER-GN \cite{Nina25}. This algorithm addresses the imbalance in outage severity within these events. 
SMOGN generates synthetic data by applying three key techniques: random under-sampling, SMOTER (Synthetic Minority Over-sampling Technique for Regression), and adding Gaussian noise. Random under-sampling involves randomly removing samples from the majority class. 
It generates new synthetic samples by interpolating rare cases. 
This is done by taking a weighted average of the target variable values between the original minority instance and its k-nearest neighbors \cite{Nina25}.

\begin{figure}[h!]
\centerline{\includegraphics[scale=0.45]{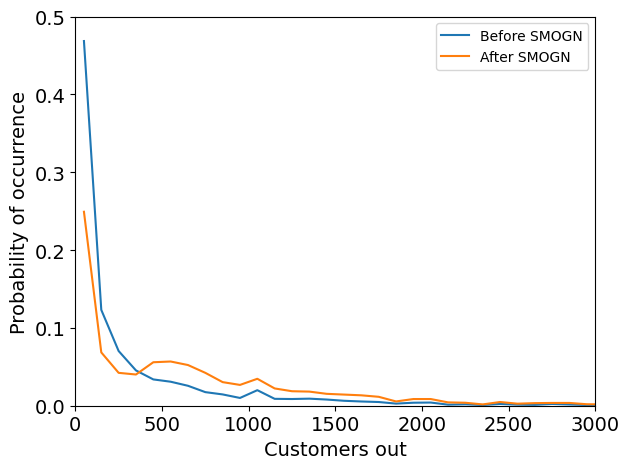}}
\caption{Probability distribution of customer outages in the EAGLE-I dataset (2014–2024), shown before and after applying SMOGN.}
\label{SMOGN}
\end{figure}

SMOGN is summarized in Algorithm \ref{alg:elsavior}: We mark a case as rare if its outage value is in the top 30\% of the target distribution. A rarity threshold of 380 customers out was used. five nearest rare dataset neighbors were considered for SMOTER interpolation within a safe region. Also, 
Gaussian noise with perturbation 2\% was added when neighbors lie outside the local safe range. Minority cases were oversampled approximately one-to-one, and the majority was undersampled to one half. 
Fig. \ref{SMOGN} shows the influence of SMOGN on the outage occurrence probability distribution for the extreme weather dataset used in the case study. The original dataset is dominated by small outages near zero, while the rebalanced dataset includes more high-consequence outages, leading to a smoother and more balanced distribution for training. After rebalancing, probability is reduced in the low-outage region and increased in the mid/high-outage region beyond the threshold. 

\subsection{Model Selection and Training}
After the input data has been rebalanced, the following ML models are trained to capture its complex patterns: 

\textbf{Random Forest (RF):}
RF for regression is an ensemble learning method that constructs multiple decision trees during training and outputs the average prediction of all trees for regression tasks. Each tree is trained on randomly selected subsets of data, reducing the correlation between trees and mitigating over-fitting. To predict customer outages number, a Random Forest Regressor was trained using weather and socio-economic features. The model is trained with 100 decision trees and evaluated on the same dataset.

\textbf{Graph Neural Network (GNN):} GNNs are deep learning architectures designed to operate on graph-structured data \cite{LiaoMPCE}. In this study, counties are represented as nodes in a spatio-temporal graph, with edges connecting geographically proximate counties (spatial edges) and linking the same county across consecutive time steps (temporal edges). Each node is characterized by weather, socio-economic, and infrastructure features. A Graph Attention Network (GAT) architecture is employed to perform message passing, where the attention mechanism adaptively weights the influence of neighboring nodes, enabling the model to focus on the most relevant spatial-temporal relationships. The model follows the procedure outlined in Algorithm \ref{alg:GNN}.

\begin{algorithm}[h!]
\caption{Spatio-Temporal Outage Prediction Using Graph Attention Network (GAT)}
\label{alg:GNN}
\begin{algorithmic}[1]
\State \textbf{Input:} $D \gets$ Dataset with hourly weather, socio-economic, spatial and extreme events data
\vspace{5pt}

\State \textbf{Target:} $y \gets$ Number of customers outage during extreme events
\vspace{5pt}

\State \textbf{Step 1: Preprocessing}
\State Normalize features $\mathcal{F}$, target $y$ using Min-Max scaling
\vspace{5pt}

\State \textbf{Step 2: Node define}
\State Assign node IDs based on (county, time)
\vspace{5pt}

\State \textbf{Step 3: Graph Construction}
\State Spatial edges: For each timestamp $t$, connect counties within 50 miles
\State Temporal edges: For each county $c$, link consecutive timestamps $t_i \rightarrow t_{i+1}$
\vspace{5pt}

\State \textbf{Step 5: Two Layers GAT Model}
\State Layer 1: $64$-dim GAT with $4$ heads + ReLU
\State Layer 2: $32$-dim GAT + ReLU
\State Output: Linear layer for scalar prediction
\vspace{5pt}

\State \textbf{Step 6: Model Training}
\For{epoch $= 1$ to $150$}
    \State Forward pass: $\hat{y} = \texttt{GAT}(G)$
\EndFor
\vspace{5pt}

\State \textbf{Step 7: Post-processing}
\State Predict scaled output $\hat{y}_{\text{scaled}}$
\State Inverse-transform to obtain actual outage: ${y} = \texttt{InverseScaler}(\hat{y}_{\text{scaled}})$

\State \textbf{Output:} Predicted customer outages $y$
\end{algorithmic}
\end{algorithm} 

\textbf{Adaptive Boosting (AdaBoost):} It is a boosting algorithm for regression which strengthens weak models by prioritizing harder-to-predict instances, refining the ensemble's accuracy with each iteration. The final prediction is a weighted sum of the predictions from all weak learners, with weights determined based on their performance. 

\textbf{Long Short-Term Memory (LSTM):} LSTM networks are a type of RNN specialized in processing sequences of data. They are particularly effective for regression tasks involving time-series data due to their ability to capture temporal dependencies over different time intervals. In this study, historical weather, socio-economic, and infrastructure features are processed through the LSTM to capture patterns over time, followed by a dense output layer for outage prediction.

\subsection{Performance Evaluation}
The following common performance metrics for regression tasks\cite{plevris2022investigation} are used for models evaluation:

\textbf{Mean Absolute Percentage Error (MAPE)}: It quantifies the average absolute deviation between predictions and observations as a percentage of the observation. 

\begin{equation}
     \mathrm{MAPE}(\%) = \left( \frac{1}{n}\sum_{i=1}^{n}
     \left|\frac{P(i)-\hat P(i)}{P(i)}\right| \quad \right)\times 100
\end{equation}

Lower values indicate better performance. For example, a MAPE of 8\% implies an average error equal to 8\%. 

\textbf{\(R^2\) Score (Coefficient of Determination)}: It quantifies the proportion of variance in the observed data that is explained by the model’s predictions. An \( R^2 \) value of 100\% indicates perfect prediction, while a value of 0 implies that the model performs no better than predicting the mean of the observed values.

\begin{equation}
R^2 = \left ( 1 - \frac{\sum_{i=1}^{n} \left(P(i) - \hat{P}(i)\right)^2}{\sum_{i=1}^{n} \left(P(i) - \bar{P}\right)^2} \right) \times 100 
\end{equation}
where \( P(i) \) is the actual outage value at instance \( i \), \( \hat{P}(i) \) is the predicted value, \( \bar{P} \) is the mean of the actual values, and \( n \) is the number of instances.

\subsection{Outage Prediction Model Formulation}

With all relevant features integrated into the dataset, the problem is formulated as predicting county-level power outages under extreme weather events. Let $y(c,t)$ denote the number of customers without power in county $c$ at time $t$ during an identified extreme-weather period. Eq. \ref{formul} shows the model formulation.
\begin{equation}
y(c,t) = f_{\theta}\!\left(
\mathbf{Y}_{c,t}^{(n)},\;
\mathbf{W}_{c,t}^{(n)},\;
\mathbf{S}_c,\;
\mathbf{I}_c
\right),
\label{formul}
\end{equation}
where $f_{\theta}(\cdot)$ denotes a data-driven mapping with parameters $\theta$ learned from historical data.

The temporal outage history is represented as
\[
\mathbf{Y}_{c,t}^{(n)} = \big[ y(c,t-1),\; y(c,t-2),\; \ldots,\; y(c,t-n) \big],
\]
and the corresponding historical weather features are given by
\[
\mathbf{W}_{c,t}^{(n)} = \big[ \mathbf{W}_{c,t},\; \mathbf{W}_{c,t-1},\; \ldots,\; \mathbf{W}_{c,t-n} \big],
\]
where $\mathbf{W}_{c,t}$ denotes the vector of contemporaneous weather variables.

The vector $\mathbf{S}_c$ contains county-level socio-economic indicators, including average household income, unemployment rate, and residential building age distribution, while $\mathbf{I}_c$ represents power infrastructure attributes.

This formulation enables the predictive models to capture nonlinear temporal dependencies in outage evolution, the influence of extreme weather dynamics, and the modulating effects of infrastructure and socio-economic vulnerability on outage severity. 

\section{Experimental Setup and Simulation Results}
The dataset spanning from 2014 to 2024 is used to train the model, while the June 6–8, 2020 flood event in Wayne County is selected for testing the performance. 
The implementation of RF and AdaBoost is done using Python Scikit-learn library \cite{N22} and LSTM is implemented using Tensorflow library \cite{N21}. The GNN model employed in this study is based on the GAT architecture using PyTorch Geometric.

Hyperparameter tuning was performed using the GridSearch tool in Python \cite{N22}, with cross-validation employed to determine the optimal input parameters for each model, as summarized in Table \ref{tableee}.
\begin{table}[h]
\centering
\caption{ Cross-Validation Hyper-Parameters Selection.}

\label{tableee}
\scalebox{1.1}{
\begin{tabular}{|c| c| c| c| c|}
\hline
 Hyper-Parameter   &       RF  & AdaBoost & GNN & LSTM\\
\hline
 Learning Rate &-& 0.001 & 0.01 & 0.001\\ \hline
 Estimators &100 & 120&-&- \\ \hline
 Hidden Units &-&-&128 $\rightarrow$ 32&128\\ \hline
 Activation &-&- &ReLU&sigmoid \\ \hline
 Optimizer &-&-&Adam&Adam \\ \hline
 Max Depth &-&-&-&-\\ \hline
 Epochs &-&-&300&100\\ \hline

\end{tabular}}
\end{table}

\begin{figure}[ht!]
\hspace{-1cm}
\centering
\includegraphics[scale=0.45]{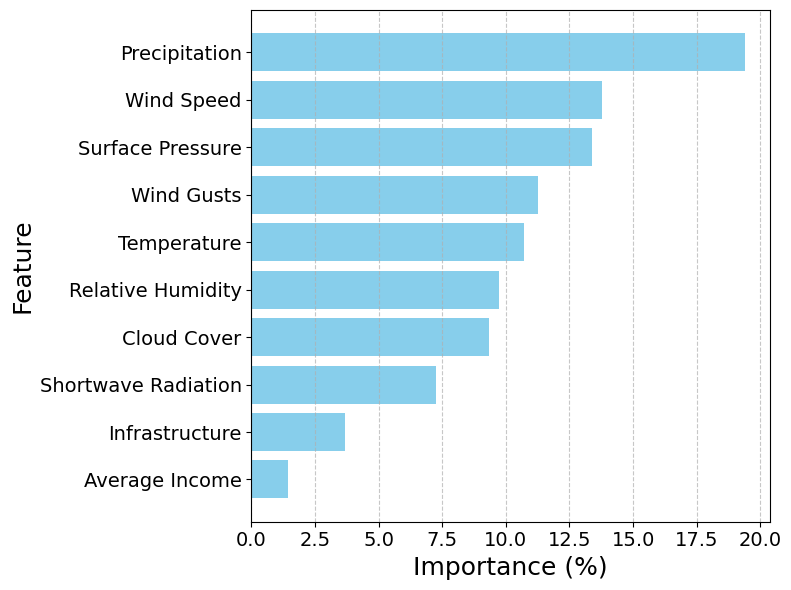}
\caption{ RF features importance across weather, infrastructure, and socioeconomic variables.}
\label{corr}
\end{figure}

Fig. \ref{corr} depicts the contribution of each feature to the model’s predictive accuracy. The analysis result showed that weather features, such as precipitation and average wind speed, have the most significant contribution to the model's performance. The analysis confirms that weather variables dominate outage prediction, with precipitation (19\%), wind speed (13.5\%), and surface pressure (13.1\%) being the top contributors. Non-weather features, including infrastructure density (3.5\%) and socioeconomic indicators such as average income (1.4\%), also exhibit contribution to the prediction.

\begin{figure}
    \centering
    \begin{subfigure}{0.65\linewidth}
        \includegraphics[width=\linewidth]{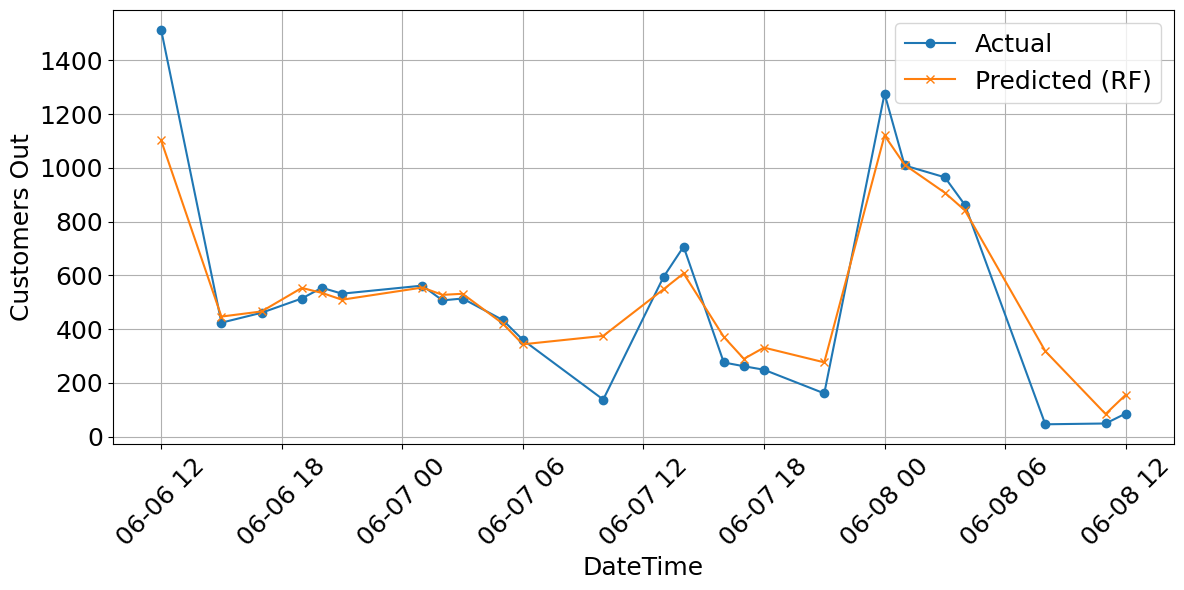}
        \caption{Random Forest prediction vs. actual outages}
        \label{fig:rf_pred}
    \end{subfigure}
    
    \begin{subfigure}{0.65\linewidth}
        \includegraphics[width=\linewidth]{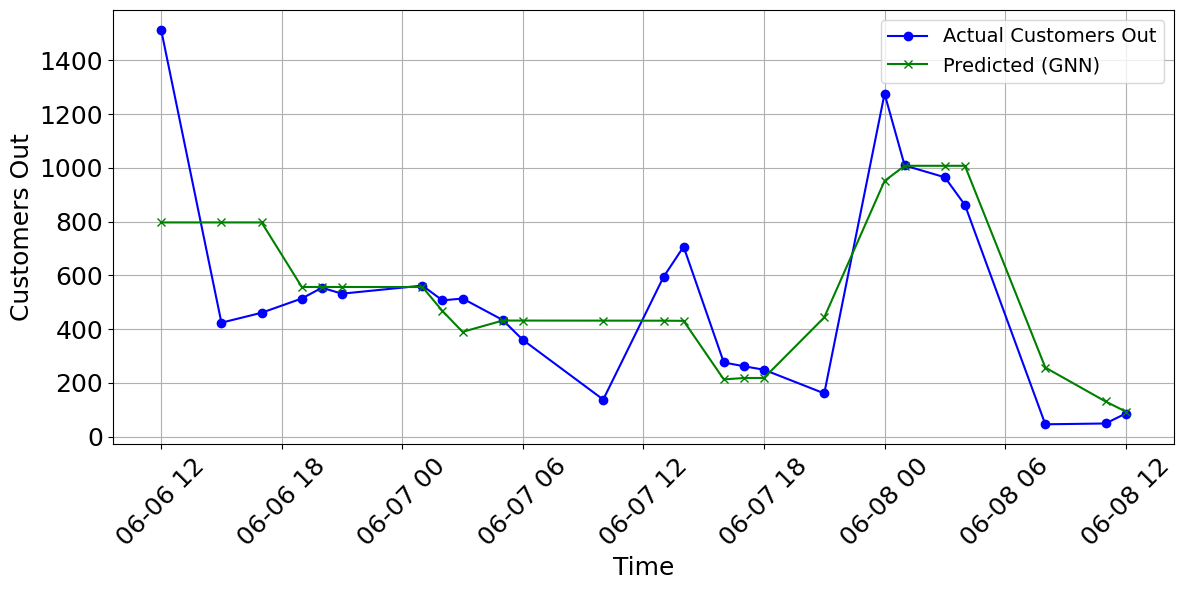}
        \caption{GNN prediction vs. actual outages}
        \label{fig:gat_pred}
    \end{subfigure}

    \begin{subfigure}{0.65\linewidth}
        \includegraphics[width=\linewidth]{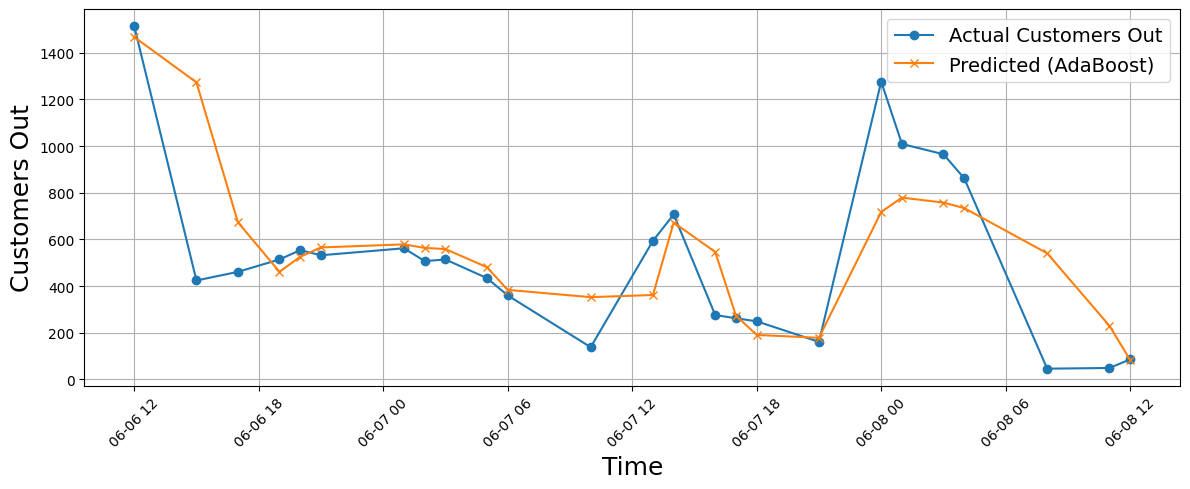}
        \caption{AdaBoost prediction vs. actual outages}
        \label{fig:adaboost_pred}
    \end{subfigure}

    \begin{subfigure}{0.65\linewidth}
        \includegraphics[width=\linewidth]{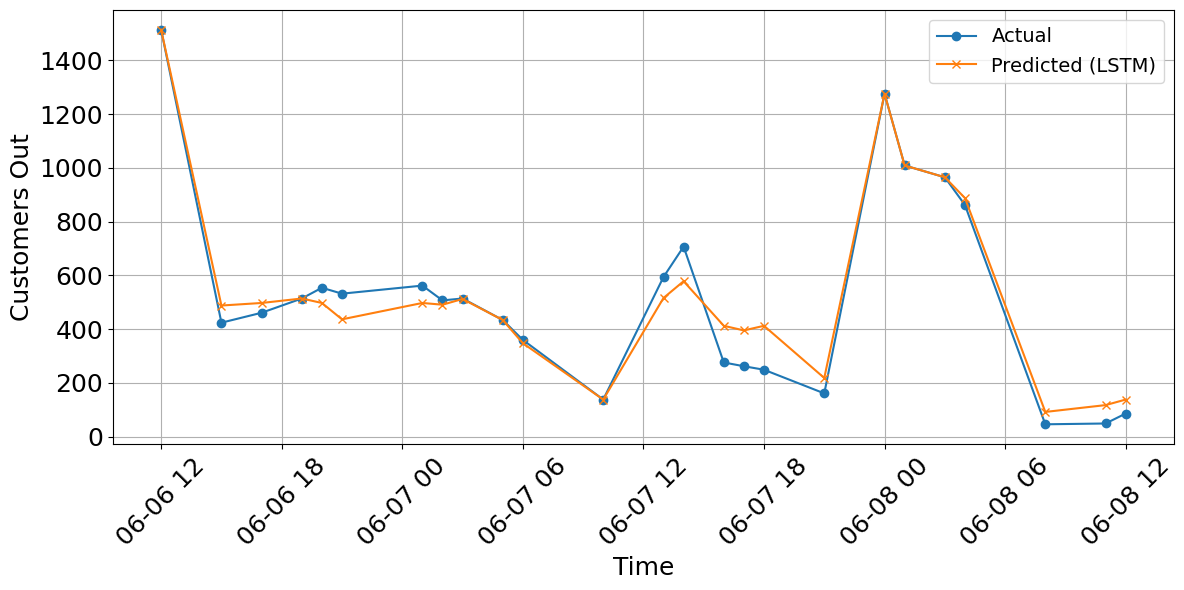}
        \caption{LSTM prediction vs. actual outages}
        \label{fig:lstm_pred}
    \end{subfigure}

    \caption{Comparison of predicted and actual outage counts for flood event in Wayne County (June 6--8, 2020).}
    \label{fig:model_predictions}
\end{figure}

Fig. \ref{fig:model_predictions} illustrates the comparison between the predicted and actual customer outage counts for the June 6–8, 2020 flood event in Wayne County, Michigan using four models: RF, GNN, AdaBoost, and LSTM. The time series plots show how well each model follows the temporal evolution and magnitude of the observed outages.
Among the four models, the LSTM provides the closest match to the actual outage pattern. This indicates that the LSTM is able to effectively exploit temporal dependencies in the sequence of weather and related features during the event window. The Random Forest model also exhibits good performance. The GNN model shows moderate agreement with the actual outages. The AdaBoost model presents the weakest performance among the four.

\begin{figure}[ht!]
\centering
\includegraphics[scale=0.4]{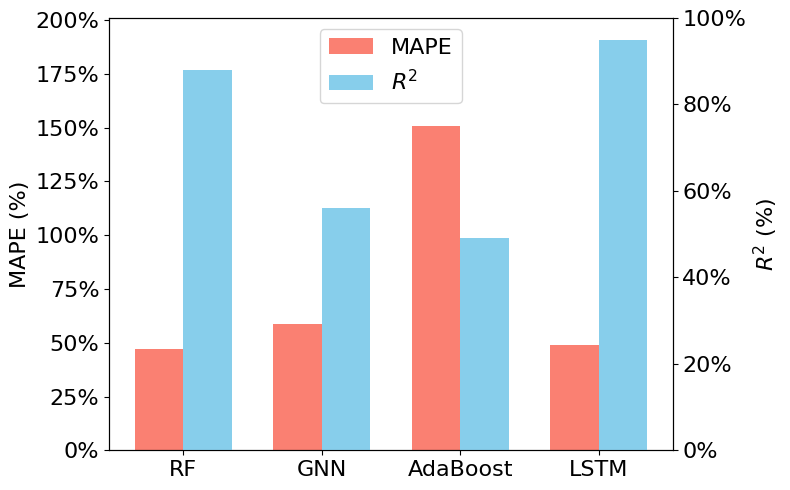}
\captionsetup{font=normalsize}
\caption{ Performance comparison between models in terms of Error (MAPE) and Accuracy ($R^2$). }
\label{performance}
\end{figure}

Fig. \ref{performance} displays the performance of each model. Fig. \ref{performance} confirming that LSTM (highest $R^2$) offers the most accurate predictions for outage events in this scenario. The RF and GNN model presents a promising deep learning-based approach with potential for further enhancement, especially for spatial-temporal generalization.

\begin{table}[h]
\centering
\caption{Comparison of Proposed Work with State-of-the-Art Studies.}
\begin{tabular}{|p{1.2cm}|p{2.5cm}|p{2.5cm}|p{2cm}|p{4.5cm}|}
\hline
\textbf{Study} & \textbf{Dataset} & \textbf{Features Used} & \textbf{Models Applied} & \textbf{Prediction Setup and Results}\\ \hline
Lee et al. \cite{N41} & EAGLE-I outage data & Historical weather & RF, KNN, XGBoost. XGBoost achieved the best performance. & The study forecasts state-level maximum outages within 12 hours of a weather alert; however, it does not offer high-resolution predictions of outages caused by extreme events. \\ \hline
Cruz et al. \cite{cruz} & EAGLE-I + NASA (weather), Census Bureau (socio-economic), FEMA Risk Index & Weather + Socio-economic & Regression, RF, XGBoost, KNN. RF outperform other models. & The model predicts next-day average outages during Hurricane Matthew (Oct 2016). It does not offer high-resolution predictions. 
\\ \hline
\textbf{Our Work (Proposed)} & EAGLE-I  + Open-Meteo (weather)+ Census (socio-economic), Infrastructure data & Weather + Socio-economic + Infrastructure + Seasonal indicators & RF, AdaBoost, LSTM, GNN. LSTM is the best performer. & The model achieves county-level validation with hourly resolution in predicting affected customers during the June 2020 Wayne County flood. \\ \hline
\end{tabular}
\label{comparison_table}
\end{table}

The performance of our proposed outage prediction framework is also compared with state-of-the-art studies in Table \ref{comparison_table}. To ensure fairness, the comparison is limited to works that employed the EAGLE-I dataset as the primary source of historical outage records. Table \ref{comparison_table} highlights that prior studies either rely on limited feature sets or provide low-resolution prediction outputs, such as maximum or averages outage customers number. In contrast, our method integrates weather, socio-economic, infrastructure, and seasonal indicators to deliver accurate county-level, hourly outage predictions during extreme events.
    
Given that this model is primarily trained on data from extreme weather and socio-economic features, it is well-suited to be used for planning to enhance the resilience of power systems. Planning can be done through simulating the system’s performance across a range of severe weather scenarios. By running these simulations, planners can assess vulnerabilities, analyze system responses, and make data-driven decisions on where to allocate resources effectively to enhance grid resilience under severe conditions.

\section{Conclusion}
In this paper, we presented a novel machine learning–based approach for predicting power outages caused by HILP events. The proposed method integrated historical outage data from the EAGLE-I dataset, weather data from Open-Meteo, as well as infrastructure and socio-economic information to construct predictive models.
We addressed the challenges of limited data availability and missing records for HILP events by applying KNN and SMOGN methods. 
Several machine learning models, including RF, GNN, AdaBoost, and LSTM networks, were compared. These models were trained and tested across scenarios with varying complexity.  

The experimental evaluation on 20 Michigan counties demonstrated that the LSTM model consistently outperformed RF, GNN and AdaBoost. Our analysis indicates that weather-related variables—particularly precipitation, wind speed, and surface pressure—are the primary drivers of outages, while infrastructure and socio-economic factors exert secondary influences.  

The model can be adapted to other regions by incorporating localized data, including region-specific weather patterns and infrastructure conditions and local social-economic information. Similarly, extending the model to handle other types of extreme events (e.g., wildfires and storms) could be achieved by adding relevant features and retraining the model on new data.

\section*{Acknowledgment}

This paper is partially funded by the Department of Energy, Solar Energy Technologies Office (SETO) Renewables Advancing Community Energy Resilience (RACER) program under Award Number DE-EE0010413. The content of this material represents the views of the authors only and is not endorsed by or necessarily reflective of the views of the Department of Energy.


\end{document}